\documentclass[english,letter]{IEEEtran}
\usepackage{mathrsfs}
\usepackage{amsfonts}
\usepackage{graphicx,cite,epsfig,amssymb,amsmath}
\usepackage{color,xcolor}
\usepackage{pifont}
\usepackage{stmaryrd}
\usepackage{setspace}
\usepackage{subfigure}
\usepackage{cite}
\usepackage{array}
\usepackage{float}
\usepackage{multirow}
\usepackage{algorithmic,algorithm}
\usepackage{epstopdf}
\usepackage{mathtools}
\floatstyle{ruled}

\providecommand{\algorithmname}{Algorithm}
\floatname{algorithm}{\protect\algorithmname}

\begin{document}
\title{A New Distributed Method for Training Generative Adversarial Networks}

\author{Jinke Ren, Chonghe Liu, Guanding Yu, and Dongning Guo

\thanks{J. Ren, C. Liu, and G. Yu are with the College of Information Science and Electronic Engineering, Zhejiang University, Hangzhou 310027 (e-mail: \{renjinke, liuchonghe, yuguanding\}@zju.edu.cn).}
\thanks{Dongning Guo is with the Department of Electrical and Computer Engineering, Northwestern University, Evanston, IL 60208 USA (e-mail: dGuo@northwestern.edu). }
\thanks{D.~Guo's work supported by the National Science Foundation under Grants No.~CCF-1910168, No.~CNS-2003098, AST-2037838, and AST-2037852 as well as a gift from Intel Incorporation.}
} \IEEEaftertitletext{\vspace{-0.75\baselineskip}}

\maketitle
\begin{abstract}
Generative adversarial networks (GANs) are emerging machine learning models for generating synthesized data similar to real data by jointly training a generator and a discriminator. In many applications, data and computational resources are distributed over many devices, so centralized computation with all data in one location is infeasible due to privacy and/or communication constraints. This paper proposes a new framework for training GANs in a distributed fashion: Each device computes a local discriminator using local data; a single server aggregates their results and computes a global GAN. Specifically, in each iteration, the server sends the global GAN to the devices, which then update their local discriminators; the devices send their results to the server, which then computes their average as the global discriminator and updates the global generator accordingly. Two different update schedules are designed with different levels of parallelism between the devices and the server. Numerical results obtained using three popular datasets demonstrate that the proposed framework can outperform a state-of-the-art framework in terms of convergence speed.
\end{abstract}

\section{Introduction}
The troika of big data, learning algorithms, and computing capability has made machine learning a key technology in the current industrial revolution \cite{6G_Chen}. Oftentimes, the tension between data collection by distributed devices and high communication cost make traditional ``cloud-based" learning paradigm unsuitable in many privacy-sensitive and resource-limited scenarios \cite{FL_magazine}. Distributed learning is a  promising paradigm that enables multiple devices to collaboratively train a machine learning model without having to share their private data \cite{Distributed_learning_survey}.

Generative adversarial networks (GANs) are popular unsupervised machine learning models that aim to approximate the statistics of a large amount of data by jointly training two neural networks, namely, a \textit{generator} and a \textit{discriminator}   \cite{Goodfellow}. GANs are often difficult to train since it is a nonconvex-nonconcave minimax problem and may suffer from mode collapse and discriminator winning issues. On the other hand, training data are often collected by network edge devices, where the data in an individual device may not be sufficient. In addition, data privacy concerns may rule out the possibility of uploading data to a central server. As an example,  a number of healthcare institutions each possesses a very limited number of diagnostic images, which are not allowed to be shared directly with other institutions due to privacy protocols. These considerations give rise to the need for a new framework to train a GAN model in a distributed and yet collaborative manner.  This will allow every institution to benefit from all data without unwarranted data exchanges \cite{BGAN}.

To effectively train GANs in distributed systems requires to jointly design communication and computation algorithms \cite{MDGAN,DecentralizedGAN,Liumingrui,FedGAN}. The authors of \cite{MDGAN} proposed a multi-discriminator framework that trains one discriminator at each device along with one generator at a single server by communicating the generated data. Further, a forgiver-first update framework was developed in \cite{DecentralizedGAN}, where the generator is updated using the output of the most forgiving discriminators. In \cite{Liumingrui}, a decentralized parallel optimistic stochastic gradient algorithm was proposed to reduce the communication overhead for training GANs and its non-asymptotic convergence was also established. Moreover, a prominent framework based on federated learning, namely FedGAN, was developed in \cite{FedGAN} by iteratively aggregating locally-trained generators and discriminators at a central server.

In this paper, we introduce a new distributed framework that enables multiple devices and a single server to collaborate in training a GAN: Each device trains a local discriminator and uploads it to the server; the server aggregates their results, computes a global GAN, and broadcasts it to all devices.  This is in contrast to FedGAN, in which each device computes both a local generator and a local discriminator, whereas the server only does model averaging.  Hence a device's computation complexity in each iteration is nearly halved in the proposed framework. Moreover, the communication overhead is also reduced because the devices in the proposed framework only upload the local discriminators in lieu of the local GANs. To boost practical implementation, we design two learning update schedules of the generator and the discriminators. We also simulate a 10-device scenario with three popular datasets, i.e., CelebA, CIFAR-10, and RSNA Pneumonia. Numerical results demonstrate that the proposed framework can achieve faster convergence speed than the FedGAN framework. We attribute the success in part to the design that allows the devices to focus on training their discriminators that depend crucially on local data, and allows the server to focus on training the generator against all discriminators in an average sense.

The rest of this letter is organized as follows. Section II introduces the system model and proposes the distributed framework. Section III develops the two learning update schedules. Numerical results are provided in Section IV. Section V concludes the letter.
\section{System Model and Distributed Framework}
\subsection{System Model}
%\begin{figure}[!htp]
%\begin{center}
%	\includegraphics[width=3.2in]{System_Model.eps}\\
%	\caption{Distributed GAN system model.}	
%	\label{Distributed_GAN_system_model}
%\end{center}\label{System}
%\end{figure}
%As depicted in Fig. 1, 
We consider a distributed system consisting of a single server and $K$ devices whose indexes form a set $\mathcal{K} = \{1,\cdots,K\}$. Device $k$ has a private dataset with $n_k$ data points, denoted as $\mathscr{X}_k=\left({\bf{x}}_{k}^1,\cdots,{\bf{x}}_{k}^{n_k}\right)$. A shared GAN model, including a generator and a discriminator, is deployed at all devices and the server, which needs to be collaboratively trained. Let $G\left(\theta,{\bf{z}}\right)$ denote the generator, where $\theta$ denotes its parameters and $\bf{z}$ denotes its input (noise). Meanwhile, let $D\left(\varphi, {\bf{d}}\right)$ denote the discriminator, where $\varphi$ denotes its parameters and $\bf{d}$ denotes its input (either real or synthesized data). The generator outputs synthesized data and the discriminator outputs an estimate of the probability that the input data is real. In the training duration, the discriminator strives to make $D\left(\varphi, G\left(\theta,{\bf{z}}\right)\right)$ approach $0$ while the generator strives to make the same quantity approach $1$. The training goal is to reach a Nash equilibrium that $G\left(\theta,{\bf{z}}\right)$  is drawn from the same distribution as the training data and $D\left(\varphi, {\bf{d}}\right)=\frac{1}{2}$ for all ${\bf{d}} \in \mathscr{X}_k, \forall k \in \mathcal{K}$.
\subsection{Distributed Framework}
To exploit distributed computational resources in the devices and the server without compromising data privacy, we jointly train a generator at the server, called \textit{global generator} and a discriminator at each device, called \textit{local discriminator}. A \textit{global discriminator} is eventually produced by taking the average of the local discriminators. We use the mini-batch stochastic gradient descent (SGD) method to update both models. The goal is to obtain a desired GAN model by iteratively exchanging the model parameters between the devices and the server. For convenience, we define two gradient functions following \cite{Goodfellow}:
\begin{align}
       & {\bf{g}_{\theta}}\left( \theta,\varphi, {\bf{z}}\right) \!= \! \nabla_{\theta} \log \left( 1 \!-\! D\left(\varphi, G\left(\theta,{\bf{z}}\right)\right)\right),\\
       &{\bf{g}_{\varphi}}\left( \theta,\varphi, {\bf{z}},{\bf{x}}\right) \!=\! \nabla_{\varphi} \left[\log D\left(\varphi, {\bf{x}}\right) \!+\! \log \left(1 \!-\! D\left(\varphi, G\left(\theta,{\bf{z}}\right)\right)\right)\right].
\end{align}
where $\nabla$ is the gradient operator.

We introduce a distributed learning framework with three component algorithms described in Algorithms 1 to 3. The key steps are summarized as follows: Device $k$ updates its local discriminator $\varphi_k$ by performing an $n_d$-step mini-batch SGD algorithm (Algorithm 1) and sends the updated parameters to the server. The server aggregates the received local discriminators using Algorithm 2 and updates the global generator by performing an $n_g$-step mini-batch SGD algorithm (Algorithm 3). We note that due to communication resource and/or limitations  by design, the server may schedule a subset of devices to participate in each iteration, whose indexes form a set $\mathcal{S} \subseteq \mathcal{K}$  via a pre-determined scheduling method, such as round-robin or proportional fair scheduling.
\begin{algorithm}[H]
	\caption{Device $k$'s update}
	{\normalsize
	\begin{algorithmic}[1]
         \STATE {\bf{Input:}} $\theta,\varphi$.
         \STATE Pick the sample size $m_k$ and the learning rate $\eta_d$.
         \STATE $\varphi_{k,0} \leftarrow \varphi$.
         \STATE \textbf{for} $j = 1,\cdots,n_d$ \textbf{do}
         \STATE ~~~~Take samples $\left( {\bf{z}}_{k,j,i},{\bf{x}}_{k,j,i}\right)_{i=1,\cdots,m_k}$.
         \STATE ~~~~Compute
                 \begin{equation}
                      \varphi_{k,j} \leftarrow \varphi_{k,j-1} + \eta_d \frac{1}{m_k} \sum_{i=1}^{m_k} {\bf{g}_{\varphi}}\left(\theta,\varphi_{k,j-1}, {\bf{z}}_{k,j,i},{\bf{x}}_{k,j,i}\right).
                 \end{equation}
         \STATE \textbf{end for}
         \STATE {\bf{Output:} $\varphi_{k,n_d}$}.
    \end{algorithmic}}
\end{algorithm}
\begin{algorithm}[H]
	\caption{Server discriminator averaging}
	{\normalsize
	\begin{algorithmic}[1]
        \STATE {\bf{Input:}} $\left(\varphi_k,m_k\right)_{k \in \mathcal{S}}$.
        \STATE {\bf{Output:} $\varphi = \frac{1}{\sum_{k \in \mathcal{S}} m_k} \sum_{k \in \mathcal{S}} m_k \varphi_k$.}

        \end{algorithmic}}
\end{algorithm}

\begin{algorithm}[H]
	\caption{Server generator update}
	{\normalsize
	\begin{algorithmic}[1]
        \STATE {\bf{Input:}} $\theta,\varphi$.
        \STATE Pick the sample size $M$ and the learning rate $\eta_g$.
        \STATE $\theta_0 \leftarrow \theta$.
        \STATE \textbf{for} $j = 1,\cdots,n_g$ \textbf{do}
        \STATE ~~~~Take samples $\left( {\bf{z}}_{j,i}\right)_{i=1,\cdots, M}$.
        \STATE ~~~~Compute
            \begin{equation}
                \theta_{j}  \leftarrow \theta_{j-1} - \eta_g \frac{1}{M} \sum_{i=1}^{M} {\bf{g}_{\theta}}\left(\theta_{j-1},\varphi,{\bf{z}}_{j,i}\right).
            \end{equation}
        \STATE \textbf{end for}
        \STATE {\bf{Output:} $\theta_{n_g}$}.
    \end{algorithmic}}
\end{algorithm}

\section{Two Learning Update Schedules}
In this section, we propose a parallel update schedule and a serial update schedule. The former allows the server and the devices to update their models in parallel while the latter requires that the device updates precede the server update.
\subsection{The Parallel Update Schedule}
\begin{figure}[!htp]
\begin{center}
	\includegraphics[width=3.5in]{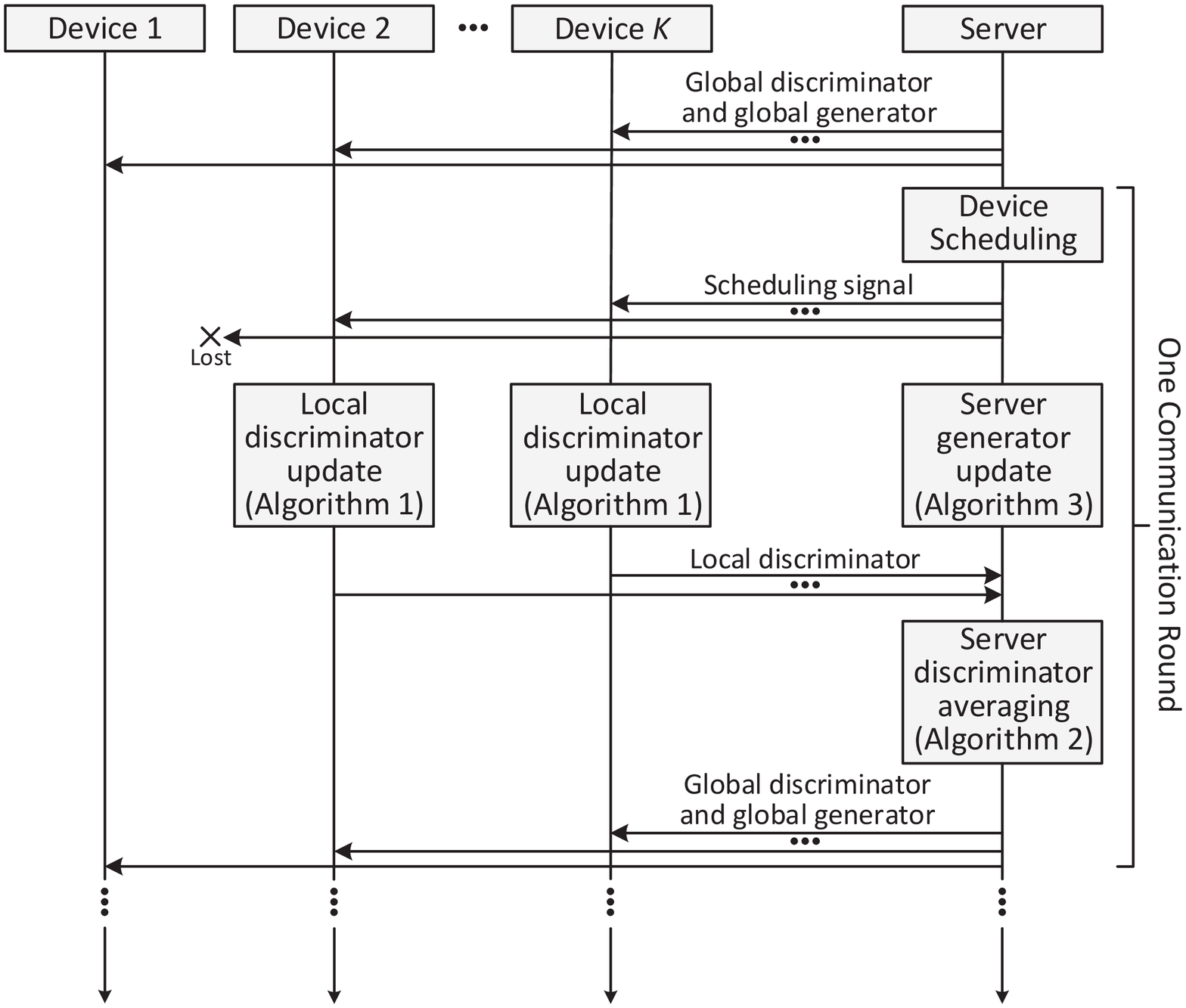}\\
	\caption{The parallel update schedule. The scheduling signal to device 1 is lost.}	
	\label{The parallel update schedule}
\end{center}
\end{figure}

In the training duration, the server and the devices can simultaneously update their models based on the same parameters in the last iteration. In this case, the inputs of Algorithm~1 and Algorithm 3 are identical. In particular, the sampled noise for global generator update should be consistent with those for local discriminator update. Therefore, we assume that the server and all devices use an identical pseudo random sequence. Specifically, the selected device $k$ shares a seed and the sampled size $m_k$ with the server either through a prior agreement or via a concurrent communication. The detailed procedure for parallel update schedule is described as follows and is also presented in Fig. \ref{The parallel update schedule}. Together, the following five steps are referred to as a \textit{communication round}.
\begin{itemize}
    \item Step 1 (Device scheduling and resource allocation):
     The server selects a subset of devices to participate in this round and allocates communication resources accordingly. 
    \item Step 2 (Local discriminator and global generator update): The scheduled devices and the server generate the noise and update their models according to Algorithm 1 and Algorithm 3, respectively.
    \item Step 3 (Local discriminator upload): The scheduled devices upload their updated discriminators to the server. 
    \item Step 4 (Server discriminator averaging):
        Upon receiving the local discriminators, the server computes the global discriminator according to Algorithm 2.
    \item Step 5 (Global discriminator and global generator broadcast): The server broadcasts the latest global discriminator and global generator to all devices.
\end{itemize}
The server and devices iterate the preceding steps until convergence.\footnote{If a device fails to receive its scheduling signal or fails to finish its local update in time, it will be excluded from the communication round.}

%We note that this learning update sequence allows the server and devices to update their models in parallel such that the computation time in one communication round can be significantly reduced.
\begin{figure}[!htp]
\begin{center}
	\includegraphics[width=3.5in]{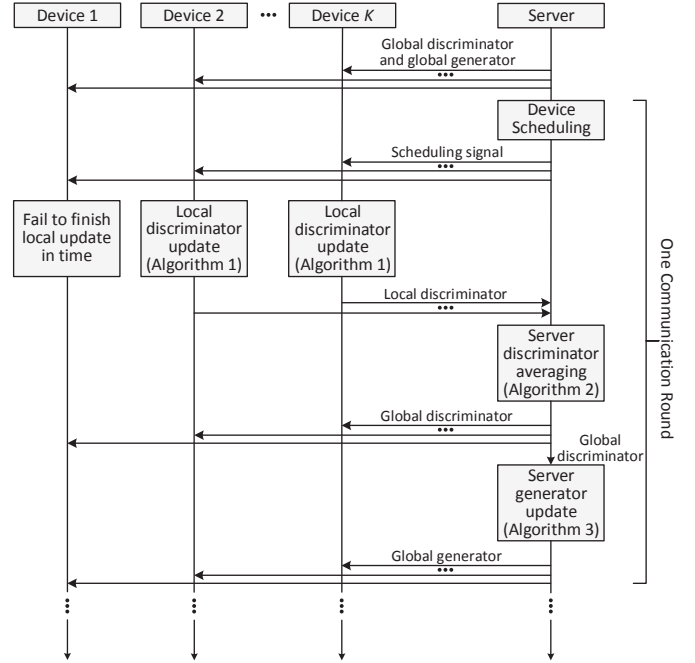}\\
	\caption{The serial update schedule. Device 1 fails to finish its local update in time.}	
	\label{The serial update schedule}
\end{center}
\end{figure}
\subsection{The Serial Update Schedule}
The serial update schedule differs from the parallel update schedule in the following manner: In the $t$-th communication round, the scheduled devices first run Algorithm 1 to update their local discriminators in parallel and send the updated model $\varphi_k^{t+1}$ to the server. The server then computes the average as the global discriminator. Thereafter, the global discriminator $\varphi^{t+1}$ is used to update the global generator, i.e.,  $\varphi^{t+1}$ is the input of Algorithm 3. Moreover, the global discriminator is broadcast to all devices once the server finishes the preceding Step 4 to reduce communication time. The detailed procedure for serial update schedule is shown in Fig. \ref{The serial update schedule}. We also note that the one-round time in serial update schedule is longer than that of parallel update schedule but it will take fewer rounds to converge according to similar stopping criteria.
\section{Numerical Results}
%In this section, we conduct experiments to test the performance of the proposed framework. 
%All codes to reproduce the results reported in this paper is available online at [github website]. % and are run on a Linux server equipped with four NVIDIA GeForce GTX 1080 Ti GPUs.
The default simulation settings are set as follows unless specified otherwise. We consider a small-cell network having a radius of $300$ m with a server located at its center servicing $K=10$ uniformly distributed devices. The path loss between each device and the server is generated by $128.1 + 37.6\log_{10}(d)$ (in dB), where $d$ is the device-to-server distance in kilometer. The channel noise power spectral density is $-174$ dBm/Hz. The transmit powers of each device and the server are $24$ dBm and $46$ dBm, respectively. The system bandwidth is $10$ MHz. The average quantitative bit number for each parameter element is $16$ bits.

We use a well-known ``DCGAN" model for experiment, where the number of parameters of its generator and discriminator are 3,576,704 and 2,765,568, respectively \cite{DCGAN}. Three popular datasets, i.e., CelebA, CIFAR-10, and RSNA Pneumonia are employed, where each dataset is randomly partitioned and assigned to the devices with equal size. The number of local iterations are set as $n_d=n_g=5$. The sampled size is $m_k=128$. In addition, we consider the Fr{\'{e}}chet inception distance (FID) metric to evaluate the performance, which characterizes the difference between the synthesized data distribution and the real data distribution \cite{FID}. A smaller FID value implies a better performance.
\begin{figure}[htp]
\begin{center}
	\includegraphics[width=2.6in]{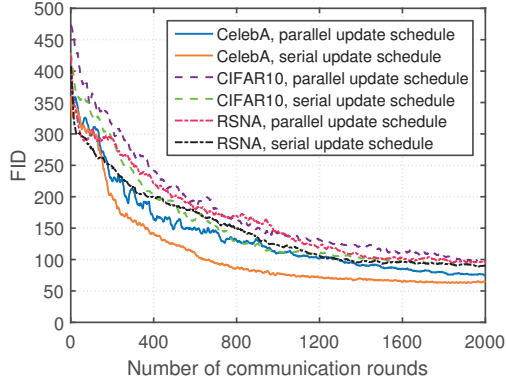}\\
	\caption{Learning performance with three datasets.}	
	\label{Generalization ability}
\end{center}
\end{figure}
\begin{figure}[htp]
\begin{center}
	\includegraphics[width=2.6in]{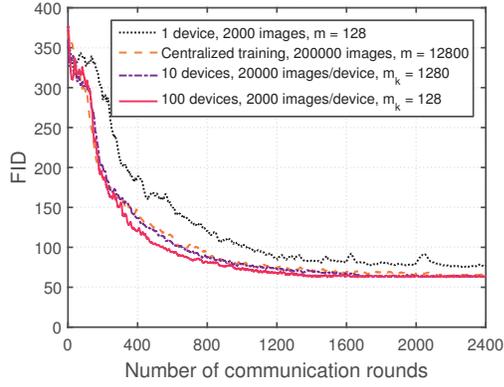}\\
	\caption{Performance comparison with different number of devices.}	
	\label{Impact of Device Number}
\end{center}
\end{figure}
\begin{figure}[htp]
\begin{center}
	\includegraphics[width=2.6in]{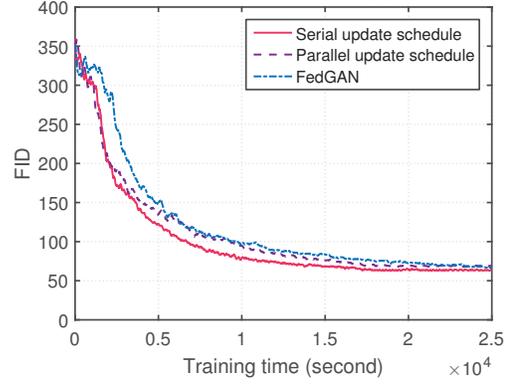}\\
	\caption{Comparison result with FedGAN.}	
	\label{Comparison result with FedGAN}
\end{center}
\end{figure}
\begin{figure}[htp]
\begin{center}
	\includegraphics[width=2.6in]{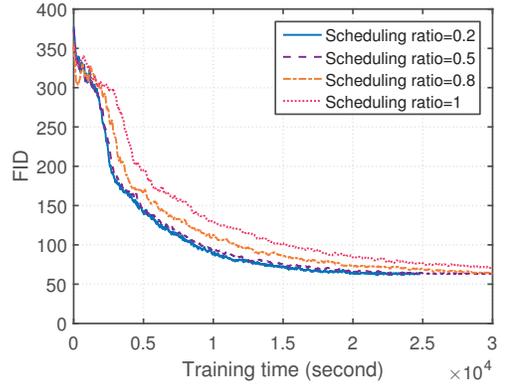}\\
	\caption{Performance with different scheduling ratios.}	
	\label{Performance with different scheduling ratios}
\end{center}
\end{figure}

Fig. \ref{Generalization ability} depicts the performance of the two proposed learning update schedules using the three datasets. We can observe that all curves converge as training proceeds, demonstrating the strong generalization ability of the proposed framework. Moreover, the serial update schedule outperforms the parallel update schedule because of the limited communication bandwidth. Since the serial update schedule needs fewer communication rounds towards convergence than the parallel update schedule, it can achieve faster convergence speed. 

Fig. \ref{Impact of Device Number} shows the performance of the proposed framework with different number of devices using the serial update schedule and the CelebA dataset. It can be observed that with the same amount of training data in each iteration, distributed training with many devices appears to converge to the same value as centralized training, but slightly faster. This maybe due to that having multiple devices make it less likely to be trapped into a local optimum during the training phase.

Fig. \ref{Comparison result with FedGAN} presents the comparison result between the proposed framework and the FedGAN framework \cite{FedGAN}. We can see that the proposed framework using serial update schedule can achieve faster convergence speed than the FedGAN framework. The reason is that the proposed framework only uploads the local discriminators whereas the FedGAN framework has to upload both the locally-trained generators and discriminators. In particular, the proposed framework using parallel update schedule achieves almost the same convergence speed as FedGAN framework because of the trade-off between one-round time and the number of rounds towards convergence.

Fig. \ref{Performance with different scheduling ratios} shows the performance of the proposed framework with different number of scheduled devices under the assumption that the uploads take variable amount of time depending on individual device's channel quality. The scheduling ratio is defined as the number of scheduled devices over the total number of devices. It can be seen from all plots that scheduling $100\%$ of the devices performs the worst among all schemes. The reason is that the channel conditions of some devices may be very bad and they become the stragglers in the training process. Scheduling $50\%$ or even just $20\%$ of the devices with the best channels achieves much better performance. This implies that there exists a trade-off between communication efficiency and learning improvement.

\section{Conclusion}
In this paper, we have proposed a new distributed framework for training GANs by communicating model parameters between multiple devices and a single server. The learning algorithms for both devices and the server have been developed and two learning update schedules have been designed for practical implementation. Simulations demonstrate the performance improvement of the proposed framework as compared with the FedGAN framework.

%\section{Acknowledgements}
%We thank Drs. Xiao Fu and Mingyi Hong for useful comments.

\end{document}